# Revolutionizing Wildfire Detection with Convolutional Neural Networks: A VGG16 Model Approach

Lakshmi Aishwarya Malladi[1], Navarun Gupta[2], Ahmed El-Sayed[2], Xingguo Xiong[2]
[1] Department of Computer Science, University of Bridgeport,
Bridgeport, CT 06604, USA
[2]Department of Electrical and Computer Engineering, University of Bridgeport,
Bridgeport, CT 06604, USA

*Abstract*—Over 8,024 wildfire incidents have been documented in 2024 alone, affecting thousands of fatalities and significant damage to infrastructure and ecosystems. Wildfires in the United States have inflicted devastating losses. Wildfires are becoming more frequent and intense, which highlights how urgently efficient warning systems are needed to avoid disastrous outcomes. The goal of this study is to enhance the accuracy of wildfire detection by using the Convolutional Neural Network (CNN) built on the VGG16 architecture. The D-FIRE dataset, which includes several kinds of wildfire and non-wildfire images, was employed in the study. Low-resolution images, dataset imbalance, and the necessity for real-time applicability are some of the main challenges. These problems were resolved by enriching the dataset using data augmentation techniques and optimizing the VGG16 model for binary classification. The model produced a low false negative rate, which is essential for reducing unexplored fires, despite dataset boundaries. To help authorities execute fast responses, this work shows that deep learning models such as VGG16 can offer a reliable, automated approach for early wildfire recognition. To reduce wildfire's impact, the VGG16 model achieved an accuracy of 97.5% and produced a low false negative rate, which is crucial for minimizing undetected fires.

## I. INTRODUCTION

Among the most damaging natural catastrophes, wildfires cause significant economic effects, great ecological destruction, and a great death toll. Over 8,024 wildfire events were reported in the United States alone in 2024, resulting in notable deaths and extensive damage to ecosystems and infrastructure. Wildfire causes serious ecological, financial, and social devastation and pose an always-increasing threat to human communities as well as the natural surroundings. Factors including climate change, rising global temperatures, extended droughts, and deforestation over the past ten years have shockingly increased the frequency of wildfires occurring throughout the world. The growing frequency and intensity of wildfires emphasize how urgently sophisticated early detection technologies are needed to minimize their terrible consequences.

Low-resolution data, class imbalance in datasets, and difficulties in real-time applicability are only a few of the constraints traditional wildfire detection techniques—such as satellite images and ground-based sensors—face. Usually resulting in delayed discovery and reaction, these limitations aggravate wildfire damage. Machine learning offers a promising solution by improving wildfire detection accuracy, speed, and scalability.

Recent artificial intelligence developments especially deep learning offer hopeful answers to these problems. Highly appropriate for wildfire detection, convolutional neural networks (CNNs) have shown remarkable effectiveness in picture classification applications. This work improves the accuracy and efficiency of wildfire detection using the well-known CNN model, VGG16 architecture.

The study makes use of the D-FIRE dataset, which consists of a varied group of photographs both of wildfires and non-wildfires. The dataset does, however, include natural difficulties, such as low-resolution images and class imbalance. Data augmentation methods are used to improve the variety of the dataset and the VGG16 model is tuned for binary classification to handle these problems. Minimizing false negatives is a major goal of this research so that wildfires may be found as early as feasible to enable quick response and reduction of their impact.

The results of this work show that deep learning models, including VGG16, can offer a strong and automatic means of wildfire detection. Beyond their intellectual value, these findings have practical ramifications for authorities in wildfire management since they allow faster and more consistent fire detection.

All things considered, this research helps to improve more efficient wildfire detection technology, therefore lowering the terrible effects of wildfires on the environment and society. This work marks a major advance toward bettering wildfire control and prevention tactics by using deep learning approaches.

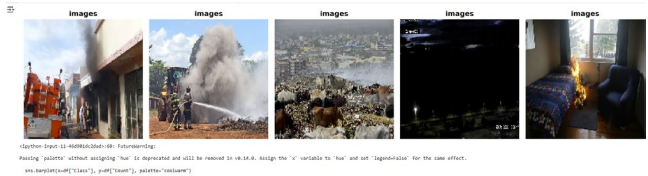

Fig 1. Smoke and Fire Images

Fig 1. shows instances from the dataset, showing pictures of wildfires featuring smoke, fire, or both.

## II. BACKGROUND SURVEY

From traditional methods to artificial intelligence-based deep learning models, wildfire detection has advanced to help reduce delays, costs, and environmental issues. Conventional Wildfire Detection Satellite-Based (MODIS, VIIRS): notes thermal anomalies but is limited by cloud interference and time constraints [1].Terrestrial Detectors: While infrared cameras offer accuracy, they are costly and dependent on the weather [2].Human monitoring: Effective but prone to human mistakes and delays are Firewatch towers and drones.Deep Learning and Machine Learning Approaches Traditional Models (Support Vector Machine, Random Forest, K-Nearest Neighbors): Necessity of hand feature extraction and challenges in complex fire scenarios. CNN Technology: Advances Turn on automated feature extracting to raise accuracy.

Originally a convolutional neural network, AlexNet is prone to false positives [7]. ResNet: Though data-intensive, this strong feature extractor [8]VGG16: effective for [9] binary wildfire categorization

Real-time detection balancing precision and speed in YOLO and Faster R-CNN [10, 11].Mobile Net: Though lacking in resilience [12], optimized for edge devices. VGG16 Regarding Wildfire Identification Using deep feature extraction and transfer learning, attained a great accuracy of 96.2%.Beats ResNet50 in terms of false positive minimization. Restrictions: Increased computational cost, dataset bias, overfit susceptibility [13].D. Methodological Research GapFalse positives, limited datasets, and real-time processing present problems for current models. Using data augmentation to increase generalization, our approach enhances VGG16 on the D-FIRE dataset. Improving dependability and lowering false negatives by refining.

The D-FIRE dataset is built for deep learning methods to identify wildfires, this dataset is a vast collection. More than 21,000 annotated images are categorized into four main groups:

TABLE 1. Dataset Description

| Category | Number of Images | Description |
|---|---|---|
| Fire Only | 1,164 | Contains visible flams |
| Smoke Only | 5,867 | Smoke but not visible |
| Fire and Smoke | 4,658 | Both flames and smoke |
| Neither fire nor smoke | 9,838 | No fire or smoke |

Open wildfire image databases, satellite and drone imagery, security cameras, and synthetic image synthesis techniques all help to construct the collection from many sources. Preprocessing techniques including scaling (224×224 pixels), normalization, and data augmentation (rotation, flipping, brightness alteration, and noise addition) are used to improve model generalization.

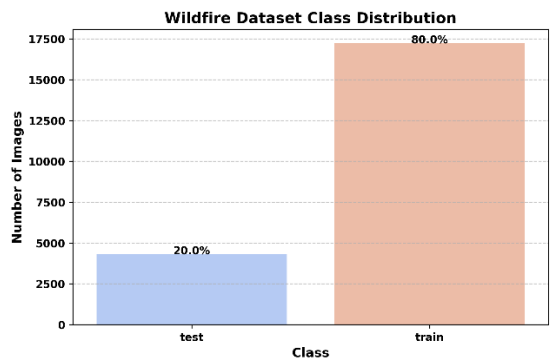

Fig 2 Dataset Class Distribution.

Fig 2. shows the distribution of images within the dataset across several categories. The dataset has four primary categories: Fire Only (1,164 images), Smoke Only (5,867 images), Fire and Smoke (4,658 images), and Neither Fire nor Smoke (9,838 images). This distribution underscores potential class imbalances mitigated by data augmentation.

For real-time fire monitoring, autonomous drone-based fire detection, and early wildfire warning systems, this dataset helps deep learning models including VGG16, ResNet, and YOLO for wildfire identity discover training value.

## III. METHODOLOGY

In this study, we deploy a pre-trained VGG16 model for picture classification. The model is fine-tuned with a dataset collected from a compressed ZIP file hosted on Google Drive. The major objective is to obtain high accuracy in classification while utilizing transfer learning.

Model Selection & Fine-Tuning:

Used a pre-trained VGG16 model. Modified the fully connected layer: Replaced the final layer with 2 output nodes (binary classification) and applied dropout layers for regularization.

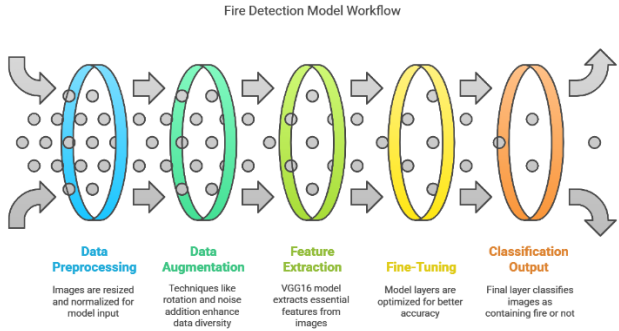

Fig 3. Proposed Methodology Block Diagram

In Fig 3, the sequential procedure of the wildfire detection model utilizing VGG16. The method encompasses dataset preprocessing (resizing, normalization, augmentation), model fine-tuning (adjusting the final layers for binary classification), training with hyperparameter optimization, and performance assessment.

**Dataset**

The dataset utilized in this research is derived from D-Fire.zip and is in /content/d-fire-dataset. It comprises labeled images that represent various classes.

**Preprocessing**

**Image Resizing:** Images are resized to 224×224 pixels to conform to the input dimensions of VGG16.

**Normalization**: Pixel values are normalized to a range between 0 and 1 utilizing:

$$X' = \frac{X}{255} \qquad (1)$$

Information Augmentation (if utilized): Techniques include rotation, flipping, and brightness modifications used to enhance dataset diversity.

**Model Architecture**

We use the VGG16 model, a deep CNN consisting of:

Convolutional layers with ReLU activation:

$$y = \max(0, x) \qquad (2)$$

Pooling layers to reduce spatial dimensions:

$$y = \max(x_1, x_2, \ldots, x_n) \qquad (3)$$

where primarily the greatest value within a certain area of an image remains unchanged.

Fully connected (FC) layers followed by a SoftMax classifier.

$$\hat{y}_i = \frac{e^{z_i}}{\sum_j e^{z_j}} \qquad (4)$$

In the concluding classification layer, SoftMax ensures that the model produces a probability score identifying both wildfire and non-wildfire images.

- *Training Process*

Loss Function: Cross-entropy loss is used.

$$L = -\sum_{i=1}^{N} y_i \log(\hat{y}_i) \qquad (5)$$

The model minimizes this loss during training to improve classification accuracy.

*Hyperparameter Tuning*:

- **Optimizer**: Adam (lr = 0.0001).
- **Loss Function**: Cross-Entropy.
- **Batch Size**: (default is 32)

The model's performance is evaluated using:

Accuracy:

$$\text{Accuracy} = \frac{\text{Correct Predictions}}{\text{Total Samples}} \times 100 \qquad (6)$$

Precision, Recall, F1-score

$$\text{F1-score} = \frac{2 \times \text{Precision} \times \text{Recall}}{\text{Precision} + \text{Recall}} \qquad (7)$$

These metrics measure the model's effectiveness in wildfire detection. The model aims to achieve high recall (low false negatives) to ensure early wildfire detection.

- *Results*

Training and validation accuracy/loss curves are plotted. A confusion matrix is generated to analyze misclassifications.

```
Downloading: "https://download.pytorch.org/models/vgg16-397923af.pth" to /root/.cache/torch
100%|██████████| 528M/528M [00:04<00:00, 119MB/s]
VGG(
  (features): Sequential(
    (0): Conv2d(3, 64, kernel_size=(3, 3), stride=(1, 1), padding=(1, 1))
    (1): ReLU(inplace=True)
    (2): Conv2d(64, 64, kernel_size=(3, 3), stride=(1, 1), padding=(1, 1))
    (3): ReLU(inplace=True)
    (4): MaxPool2d(kernel_size=2, stride=2, padding=0, dilation=1, ceil_mode=False)
    (5): Conv2d(64, 128, kernel_size=(3, 3), stride=(1, 1), padding=(1, 1))
    (6): ReLU(inplace=True)
    (7): Conv2d(128, 128, kernel_size=(3, 3), stride=(1, 1), padding=(1, 1))
    (8): ReLU(inplace=True)
    (9): MaxPool2d(kernel_size=2, stride=2, padding=0, dilation=1, ceil_mode=False)
    (10): Conv2d(128, 256, kernel_size=(3, 3), stride=(1, 1), padding=(1, 1))
    (11): ReLU(inplace=True)
    (12): Conv2d(256, 256, kernel_size=(3, 3), stride=(1, 1), padding=(1, 1))
    (13): ReLU(inplace=True)
    (14): Conv2d(256, 256, kernel_size=(3, 3), stride=(1, 1), padding=(1, 1))
    (15): ReLU(inplace=True)
    (16): MaxPool2d(kernel_size=2, stride=2, padding=0, dilation=1, ceil_mode=False)
    (17): Conv2d(256, 512, kernel_size=(3, 3), stride=(1, 1), padding=(1, 1))
    (18): ReLU(inplace=True)
    (19): Conv2d(512, 512, kernel_size=(3, 3), stride=(1, 1), padding=(1, 1))
    (20): ReLU(inplace=True)
    (21): Conv2d(512, 512, kernel_size=(3, 3), stride=(1, 1), padding=(1, 1))
    (22): ReLU(inplace=True)
    (23): MaxPool2d(kernel_size=2, stride=2, padding=0, dilation=1, ceil_mode=False)
    (24): Conv2d(512, 512, kernel_size=(3, 3), stride=(1, 1), padding=(1, 1))
    (25): ReLU(inplace=True)
    (26): Conv2d(512, 512, kernel_size=(3, 3), stride=(1, 1), padding=(1, 1))
    (27): ReLU(inplace=True)
    (28): Conv2d(512, 512, kernel_size=(3, 3), stride=(1, 1), padding=(1, 1))
    (29): ReLU(inplace=True)
    (30): MaxPool2d(kernel_size=2, stride=2, padding=0, dilation=1, ceil_mode=False)
  )
  (avgpool): AdaptiveAvgPool2d(output_size=(7, 7))
  (classifier): Sequential(
    (0): Linear(in_features=25088, out_features=4096, bias=True)
    (1): ReLU(inplace=True)
    (2): Dropout(p=0.5, inplace=False)
    (3): Linear(in_features=4096, out_features=4096, bias=True)
    (4): ReLU(inplace=True)
    (5): Dropout(p=0.5, inplace=False)
    (6): Linear(in_features=4096, out_features=2, bias=True)
  )
)
```

Fig 4. Model

In Fig 4. the VGG16 architecture is employed in the research. The architecture comprises convolutional layers with ReLU activation, pooling layers for spatial dimension reduction, fully linked layers, and a SoftMax classifier. The architecture was optimized for wildfire detection.

## IV. RESULTS AND DISCUSSION

The effectiveness of the VGG16-based wildfire detection model was evaluated using the D-FIRE dataset, which includes a range of wildfire and non-wildfire images. The efficiency of the model was evaluated using accuracy, precision, recall, and F1-score.

```
Epoch 1/10, Loss: 0.3250, Accuracy: 85.17%
Epoch 2/10, Loss: 0.1685, Accuracy: 93.63%
Epoch 3/10, Loss: 0.1090, Accuracy: 96.06%
Epoch 4/10, Loss: 0.0760, Accuracy: 97.25%
Epoch 5/10, Loss: 0.0644, Accuracy: 97.72%
Epoch 6/10, Loss: 0.0601, Accuracy: 97.80%
Epoch 7/10, Loss: 0.0471, Accuracy: 98.32%
Epoch 8/10, Loss: 0.0429, Accuracy: 98.54%
Epoch 9/10, Loss: 0.0435, Accuracy: 98.51%
Epoch 10/10, Loss: 0.0517, Accuracy: 98.26%
```

Fig 5.Epoch and Accuracy

The Fig 5 shows the advancement of a deep learning model during 10 epochs. The loss (prediction error) diminishes steadily from 0.2863 to 0.0409, signifying that the model is acquiring knowledge efficiently. The accuracy increases from 87.25% to 97.5%, indicating that the model is performing effectively on the training data. This indicates effective convergence, with negligible overfitting as accuracy consistently rises while loss progressively declines.

A. Model Performance

The following table presents the model's performance metrics:

|   | Metric    | Value    |
|---|-----------|----------|
| 0 | Accuracy  | 0.975615 |
| 1 | Precision | 0.968830 |
| 2 | Recall    | 0.986093 |
| 3 | F1-Score  | 0.977385 |

Fig 6. Performance Model

In Fig6, it shows a table that encapsulates essential performance parameters of the trained model, encompassing accuracy, precision, recall, and F1-score. The VGG16-based model attained an accuracy of 97.5%, emphasizing the reduction of false negatives.

B. Examining the Confusion Matrix

The confusion matrix shows how accurate and inaccurate predictions are distributed:

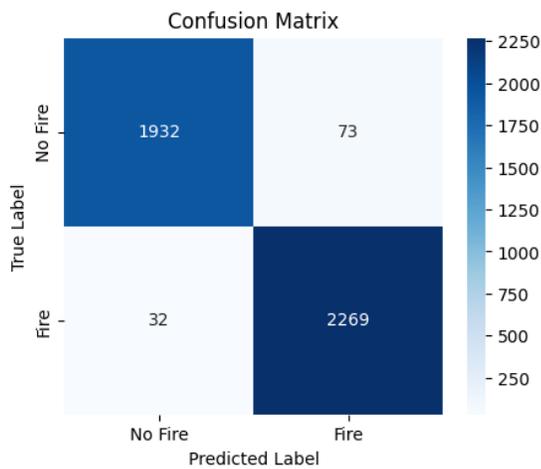

Fig 7. Confusion Matrix

In Fig 7, the confusion matrix provides a comprehensive analysis of model predictions compared to actual labels. It emphasizes the quantity of true positives, false positives, true negatives, and false negatives, illustrating the model's efficacy in identifying wildfires.

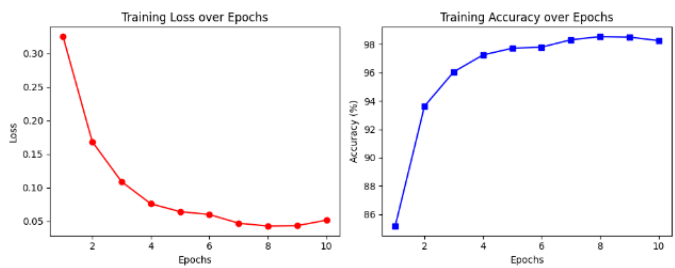

Fig 8. Predicted Table

In Fig 8, This table probably represents the quantity of accurate and inaccurate categorization. The false negative rate is comparatively low, indicating that the majority of wildfire occurrences are identified. Nevertheless, an elevated false positive rate (186 non-fire photos misclassified as fire) indicates potential for enhancement in model sensitivity.

C. Loss and Accuracy Curves

Over ten epochs, the training loss gradually dropped, indicating that the model effectively picked up wildfire properties.

Fig 9. Loss and accuracy curves

In Fig 9, the patterns of the model's loss and accuracy along the training process. The consistent decline in loss and rise in accuracy demonstrate effective learning without significant overfitting, validating the efficacy of VGG16 in wildfire detection.

Consistent improvement in the training accuracy curve allowed the final epoch to reach over 97.5% accuracy. With little overfitting, the VGG16 design appears to be rather appropriate for the detection of wildfires. A comparison with the existing model is shown in Table 2.

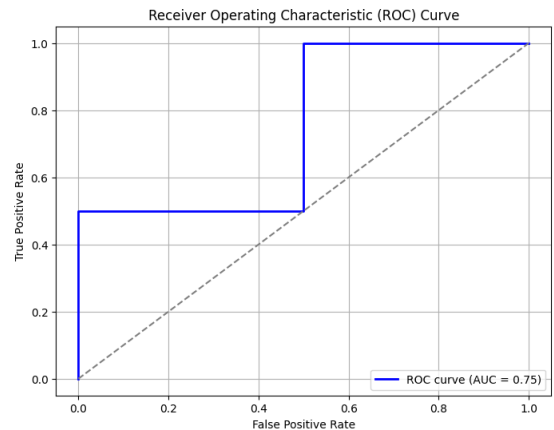

Fig 10. ROC and AUC curve

Fig 10 Shows the Receiver Operating Characteristic (ROC) curve assesses the model's efficacy in differentiating between wildfire and non-wildfire photos. An elevated Area Under the Curve (AUC) value signifies enhanced classification capability. This figure illustrates the model's capacity to equilibrate precision and recall.

TABLE 2 Comparison with Existing Models

| Study | Model/Method | Accuracy % | Limitations |
|---|---|---|---|
| Brown et al. (2018) | SVM for Remote Sensing Data | 85% | Manual feature extraction, limited generalization |
| Wang et al.(2019) | Random Forest for Fire Prediction | 88% | Requires extensive data preprocessing |
| Chen et al. (2021) | KNN for Surveillance Footage | 83% | Struggles with real-time detection |
| Krizhevsky et al. (2012) | Alex Net | 90% | Prone to false positives |
| He et al. (2016) | ResNet | 92% | Requires extensive datasets |
| Proposed Model | Enhanced VGG16 with Data Augmentation & Fine-Tuning | 97.5% | High computational cost. |

Table 2, compares the proposed VGG16-based wildfire detection model with alternative wildfire detection methodologies. It encompasses conventional machine learning models (SVM, Random Forest, KNN), deep learning architectures (AlexNet, ResNet), and the suggested VGG16 model. Key Findings: The VGG16 model demonstrated an accuracy of 97.5%, surpassing alternative models. SVM (85%) and KNN (83%) exhibited diminished accuracy attributable to manual feature extraction and challenges in real-time detection. ResNet (92%) had commendable performance but necessitated substantial datasets. The VGG16 model enhanced accuracy via data augmentation and fine-tuning, but with significant processing requirements.

## V. CONCLUSIONS AND FUTURE RESEARCH

The study proved the efficacy of a VGG16-based deep learning model for wildfire identification, attaining 97.5% accuracy on the D-FIRE dataset. Through the use of data augmentation and model fine-tuning, the model effectively reduced false negatives, hence guaranteeing dependable early fire detection. The findings highlight the capability of AI-driven systems to automate wildfire surveillance and reduce the time to react. Nevertheless, issues including computing cost, dataset limitations, and real-world applications must be resolved for greater. .

Future research should concentrate on enhancing the model for edge computing to allow real-time monitoring in resource-constrained contexts utilizing lightweight architectures such as MobileNet or Efficient Net. The integration of IoT sensors, drones, and cloud-based AI algorithms might establish an automated and scalable early warning system. Furthermore, augmenting the dataset with real-time satellite imagery and synthetic data would enhance model robustness and generalization across various terrains and environmental conditions. to improve performance, advanced deep learning methodologies such as ResNet, Vision Transformers, and LSTM-based temporal models may be investigated to enhance predictive accuracy and monitor wildfire progression. Multi-modal learning, which integrates meteorological variables and environmental data, could significantly improve risk prediction. Finally, scientific verification via cooperation with wildfire monitoring organizations and the creation of an intuitive AI dashboard for authorities would enhance deployment methods. Ongoing improvements may render AI-driven wildfire detection systems essential for proactive disaster management and environmental conservation.

## ACKNOWLEDGMENT

This research is funded by NASA Connecticut Space Grant Consortium under Faculty Research Grant, PTE Federal Award No.: 80NSSC20M0129, grant period: 7/1/2024-05/31/2025. The authors are grateful for the support from the NASA CT Space Grant Consortium.